\documentclass{article}
\usepackage{ijcai22}

\usepackage{times}
\usepackage{soul}
\usepackage{url}
\usepackage[hidelinks]{hyperref}
\usepackage[utf8]{inputenc}
\usepackage[small]{caption}
\usepackage{graphicx}
\usepackage{amsmath}
\usepackage{amsthm}
\usepackage{booktabs}
\usepackage{algorithm}
\usepackage{algorithm}
\usepackage{algpseudocode}
\usepackage{wrapfig}
\usepackage{subcaption}
\usepackage{lipsum}
\usepackage{color}
\title{Self-Predictive Dynamics for Generalization of Vision-based\\ Reinforcement Learning}

\author{
    Kyungsoo Kim
    \and
    Jeongsoo Ha\And
    Yusung Kim\thanks{Corresponding author}\\
\affiliations
    Sungkyunkwan University\\
\emails
    \{unigary, hjg1210\}@g.skku.edu, yskim525@skku.edu
}

\begin{document}

\maketitle

\begin{abstract}
  Vision-based reinforcement learning requires efficient and robust representations of image-based observations, especially when the images contain distracting (task-irrelevant) elements such as shadows, clouds, and light. It becomes more important if those distractions are not exposed during training. We design a Self-Predictive Dynamics (SPD) method to extract task-relevant features efficiently, even in unseen observations after training. SPD uses weak and strong augmentations in parallel, and learns representations by predicting inverse and forward transitions across the two-way augmented versions. In a set of MuJoCo visual control tasks and an autonomous driving task (CARLA), SPD outperforms previous studies in complex observations, and significantly improves the generalization performance for unseen observations. Our code is available at \url{https://github.com/unigary/SPD}.
  
\end{abstract}

\section{Introduction}
Vision-based reinforcement learning (RL)~\cite{hafner2019learning,srinivas2020curl,zhang2020learning} has been studied to learn optimal control using high dimensional image inputs. The demand for vision-based RL has continued to grow as more attempts are made to apply RL to real-world applications such as robotics and autonomous driving, which primarily use image data. However, to achieve this, vision-based RL must address two fundamental problems; data efficiency and generalization. Data efficiency refers to how quickly optimal control of a task can be learned using fewer experience samples. Learning control from high dimensional images such as raw pixels inevitably increases the learning difficulty. In particular, if the images contain task-irrelevant information (clouds, shadows, and light etc.), this unnecessary information interferes with learning optimal control. The more complex the observation, the worse this problem is. 
In terms of generalization, task-irrelevant information may vary depending on the time and location of the actual tests. If those distracting elements are not exposed during training, control performance could be severely degraded. Some prior works present that using relatively weak data augmentations can improve data efficiency rather than using strong augmentations~\cite{laskin2020reinforcement}. However, we found that it is not sufficient if the observed characteristics at the time of testing differ from those at the time of training as shown in Table~\ref{table_dmc_result}.


In this work, we design Self-Predictive Dynamics (SPD) as a method of self-supervised learning suitable for vision-based RL. Our method introduce two-way data augmentations which apply both weak and strong augmentation techniques for the same observation. First, we use a discriminator to distinguish between two-way augmented observations, while our encoder learns to fool the discriminator. It helps that our encoder to capture invariant features from the different-level augmented versions.
Second, SPD infers actually conducted actions between successive (latent) states across two-way augmentations. The inferred actions are used to predict the identical future states from (two-way augmented) current states. By accurately understanding dynamics chaining (from inverse to forward dynamics), 
SPD can learn optimal control policies more efficiently in complex visual environments, and shows excellent generalization performance especially for unseen observations.
\begin{figure*}[t]
\begin{center}
\includegraphics[width=0.90\textwidth]{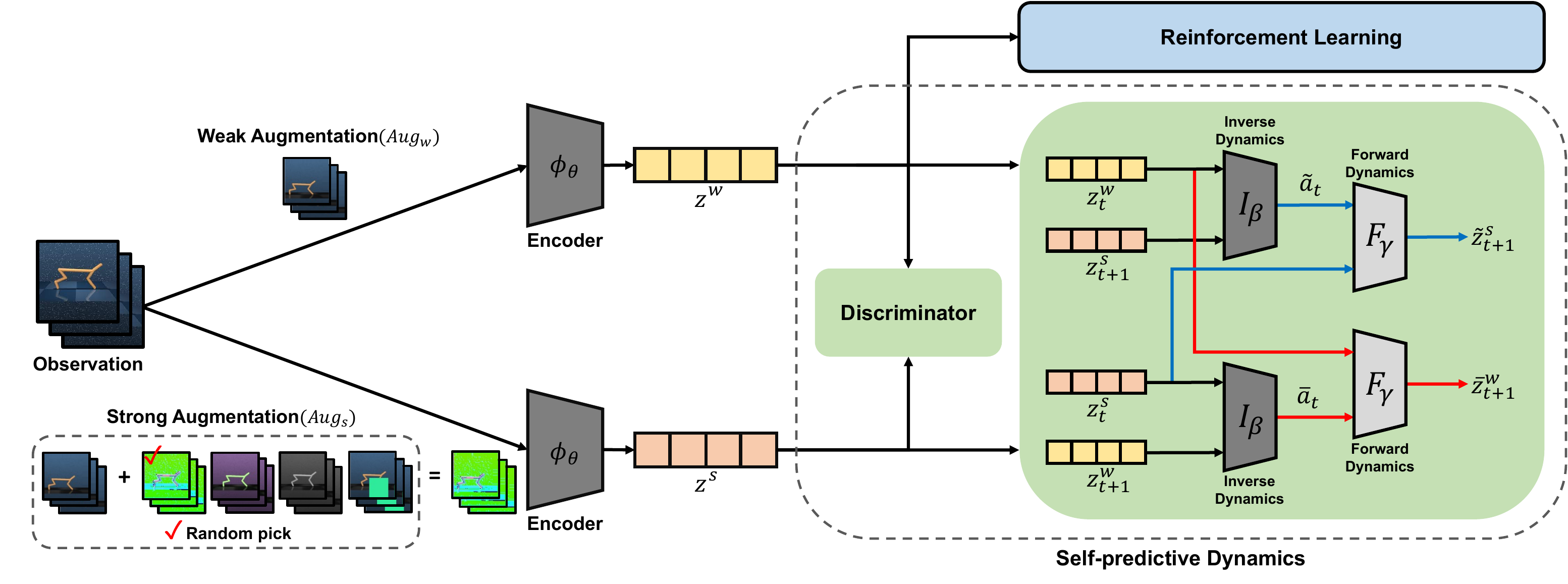}
\end{center}
\vspace{-0.3cm}
\caption{Our Framework Overview: we use a shared encoder for RL and Self-Predictive Dynamics (SPD). An observation is augmented in two ways; $Aug_w$ uses Random-Shift only, and $Aug_s$ uses Random-Shift and other randomly chosen augmentation method. The encoded latent state $z^w$ is used to train an RL algorithm, and both $z^w$ and $z^s$ are passed to SPD.}
\label{fig:overview}
\end{figure*}
For evaluation, we used a set of continuous control tasks (the DeepMind Control suite~\cite{tassa2018deepmind}) with distracting elements backgrounds as proposed in~\cite{zhang2020learning}. Compared to prior studies, SPD efficiently learns a control policy in both simple and complex observations. We also show that SPD significantly outperforms existing studies when the testing observations differ from the training observations, which means higher generalization ability. In an autonomous driving task, CARLA~\cite{dosovitskiy2017carla}, our method achieves the best performance on complex observations containing a lot of task-irrelevant information in realistic driving scenes.

The key contributions of this paper are as follows:
\begin{itemize} 
\item We introduce a Self-Predictive Dynamics (SPD) method using both weak and strong augmentations in parallel. SPD enables one-stage learning generalization without additional pre-training or fine-tuning processes.
\item SPD outperforms previous studies in complex backgrounds, and shows the best generalization performance when observed characteristics change in real-world scenarios after training.
\end{itemize}

\section{Related Works}
In vision-based RL studies, representation learning is a fundamental component for achieving an optimal control policy. 
Many studies have been conducted on improving data efficiency and generalization, mainly using data augmentation techniques and self-supervised learning methods.

\subsection{Data Efficiency in Vision-based RL} 
Some studies have introduced pixel-level reconstructions for representation learning using variational inference~\cite{yarats2019improving,lee2019stochastic}. By reconstructing the current observation accurately, it helps to extract compact representations of image observations. It has been shown that learning forward dynamics to predict the future state~\cite{schwarzer2020data,oord2018representation} can be effective for making better representations. Several RL studies proposed to use data augmentations which provide different views of the image data. RAD~\cite{laskin2020reinforcement} has shown that using data augmentations improves data efficiency without modifying RL algorithms. DrQ~\cite{kostrikov2020image} has improved data efficiency using both data augmentation methods and modified Q-functions. CURL~\cite{srinivas2020curl} has combined data augmentations and contrastive learning~\cite{chen2020simple} to learn representation more efficiently.
These studies have used relatively simple and weak such as random-crop or random-shift. Although those weak augmentations are useful to improve data efficiency in simple backgrounds, they are NOT working well for complex or unseen observations. 

\subsection{Generalization in Vision-based RL}
In vision-based control tasks, not only image observations include information not relevant to the task such as clouds, shadows, and light, but these distracting factors can change continuously over the duration of the test. Therefore, extracting invariant features relevant to the task control is a key challenge for improving generalization. 
DBC~\cite{zhang2020learning} used bisimulation metrics to provide effective downstream control by learning invariant features from the images including task-irrelevant details.
DBC shows the potential for generalization, but the performance achieved is still low. Inverse dynamics has been used as one of self-supervised auxiliary tasks in RL~\cite{pathak2017curiosity}. PAD~\cite{hansen2020self} has used inverse dynamics with weak data augmentations not only training a policy but fine-tuning to adapt the policy to new environments. 
Some recent studies have suggested the use of strong data augmentation techniques that heavily distort the image such as Color-jitter or Random-convolution~\cite{lee2019network}. Strong augmentations are known to lead to robust and generalizable representations for vision research areas, but naively applying them into RL results in sub-optimal performance~\cite{laskin2020reinforcement}. SODA~\cite{hansen2021generalization} learns representation by maximizing the mutual information between strong augmented data and non-augmented data. SECANT~\cite{fan2021secant} first learns an expert policy with weak augmentations, and imitates the expert policy with strong augmentations.

Our work suggests Self-Predictive Dynamics (SPD) across two-way (weak and strong) data augmentations in parallel. The learning process of SPD is simple and does NOT require any pre-training or fine-tuning after deployments.


\section{Self-Predictive Dynamics}
In this section, we introduce Self-Predictive Dynamics (SPD) which consists of the two-way data augmentations, discriminator and dynamics chaining. Our method does not require any changes to the underlying RL algorithm, and any RL algorithm can be used.

\subsection{Model Overview}
We design the model architecture to share represented features that feed into SPD and RL. We define encoder $\phi$, discriminator $D$, and dynamics chaining $\psi$. Our goal is to train the encoder $\phi$ to extract task-control relevant information efficiently so that the RL agent can learn the generalized optimal policy. The encoder $\phi$ is updated with the gradients of SPD and RL. The model overview is illustrated in Figure~\ref{fig:overview} and Algorithm~\ref{alg:algorithm}.

\subsection{Two-way Data Augmentations}\label{Aug}
We introduce a two-way data augmentation method. The weak and strong augmented versions are used in parallel during training.
\textbf{Random-shift}~\cite{kostrikov2020image} is used for a weak augmentation technique. It pads each side and then selects a random crop back to the original image size.
For strong augmentation techniques, we use a combination of Random-shift and a randomly chosen one among the following four techniques. \textbf{Grayscale} converts RGB images to grayscale images based on certain probabilities. \textbf{Random convolution}~\cite{lee2019network} transforms an image through a randomly initialized convolutional layer. \textbf{Color-jitter} converts RGB image to HSV image which adds noise to each channel of HSV. \textbf{Cutout-color}~\cite{cobbe2019quantifying} randomly inserts a small random color occlusion into the input image.

In Figure~\ref{fig:overview}, two-way augmentations are shown for a given observation. $Aug_{w}$ stands for a weak augmented version and $Aug_{s}$ represents a strong augmented version. In our ablation test, using multiple strong augmentation techniques together shows better performance than using a single strong augmentation in the supplementary material.


\subsection{Discriminator}
The goal of the discriminator is for the encoder to reduce the difference between the representations for the weak and strong augmented versions.
When two-way data augmentations $Aug_{w}$ and $Aug_{s}$ pass through the encoder $\phi$, it produces latent states $z^{w} = \phi(Aug_{w}(obs))$ and $z^{s} = \phi(Aug_{s}(obs))$ where $obs$ is an image observation.
For the discriminator, we use the concept of a relativistic GAN~\cite{jolicoeur2018relativistic}, which is known to be more stable and faster than a standard GAN. For $z^{w}$ and $z^{s}$, we define encoder (as a generator) and discriminator objective functions as follows, where $\sigma$ represents a sigmoid function.
\begin{equation}\label{eq1}
    J({\phi}) = -\log (\sigma(D(z^{s}) - D(z^{w}))),
\end{equation}
\begin{equation}\label{eq2}
    J({D}) = -\log (\sigma(D(z^{w}) - D(z^{s}))).
\end{equation}
$J(\phi)$ optimizes $z^{s}$ to have a higher value than $z^{w}$ in Equation~\ref{eq1}. Conversely, $J(D)$ optimizes $z^{w}$ to have a higher value than $z^{s}$ in Equation~\ref{eq2}. By alternately optimizing Equation~\ref{eq1} and Equation~\ref{eq2}, the encoder $\phi$ is updated so that the representations of $z^{w}$ and $z^{s}$ become similar.
Eventually, our discriminator helps to learn invariant features regardless of the position shifts and the changes in color and texture of the observations.

\begin{algorithm}[t]
\caption{Self-Predictive Dynamics}
\label{alg:algorithm}
\textbf{Initialize}: Encoder $\phi$, Policy $\pi$, Critic $Q$, Discriminator $D$, Dynamics chaining $\psi$, Buffer $B$.
\begin{algorithmic} 
\For{each iteration}
\For{each environment step}
\State Encode state $z_{t}=\phi(s_{t})$
\State Execute action $a_{t}=\pi(z_{t})$
\State Store transition: $B \leftarrow B \cup \left\{ s_{t},a_{t},s_{t+1},r_{t} \right\}$
\EndFor
\For{each update step}
\State Sample mini-batch: $(S, A, S', R) \sim B$
\State \textit{// Apply weak augmentation} \\ \hskip1.0cm $Z_{w}, Z'_{w} = Aug_{w}(S), Aug_{w}(S')$
\State \textit{// Apply strong augmentation} \\ \hskip1.0cm $Z_{s}, Z'_{s} = Aug_{s}(S), Aug_{s}(S')$
\State \textit{// Train self-supervisions} \\ \hskip1.0cm $E_{Z_{w}, Z'_{w}, Z_{s}, Z'_{s}, A}[J(\psi,\phi,D)]$
\State \textit{// Train RL Policy} \\ \hskip1.0cm  $E_{Z_{w}, Z'_{w}}[J(\pi)]$
\EndFor
\EndFor
\State \textbf{return} Optimal Policy $\pi$
\end{algorithmic}
\end{algorithm}

\subsection{Dynamics Chaining}
We introduce a dynamics chaining which consists of inverse dynamics and forward dynamics based on two-way data augmentations. For given sequential latent states $z^{w}_{t}$ and $z^{s}_{t+1}$, and another pair of $z^{s}_{t}$ and $z^{w}_{t+1}$, inverse dynamics $I$ infers the actions $\tilde{a}_{t} = I(z^{w}_{t}, z^{s}_{t+1})$ and $\bar{a}_{t} = I(z^{s}_{t}, z^{w}_{t+1})$. Even if the input images are augmented with different levels, the two inferred actions should be similar to each other, and should be nearly identical to the action $a_{t}$ actually performed.

The inferred actions $\tilde{a}_{t}$ and $\bar{a}_{t}$ are fed into forward dynamics $F$ along with the current latent states $z^{s}_{t}$ and $z^{w}_{t}$. $F$ predicts the next latent states as following; $\tilde{z}^{s}_{t+1} = F(z^{s}_{t}, \tilde{a}_{t})$ and $\bar{z}^{w}_{t+1} = F(z^{w}_{t}, \bar{a}_{t})$.
$\tilde{z}^{s}_{t+1}$ and $\bar{z}^{w}_{t+1}$ are predicted across two-way augmented versions in parallel, they should be identical to $z^{s}_{t+1}$ and $z^{w}_{t+1}$.
This dynamics chaining allows our encoder to learn more powerful representations by using both dynamics knowledge inferred across two-way augmented observations.

The inverse dynamics objective function Equation~\ref{eq3} is defined as the mean squared error between actual action and inferred action.
\begin{equation}\label{eq3}
    J({I}) = {\frac{(I(z^{w}_{t}, z^{s}_{t+1})-a_{t})^2 + (I(z^{s}_{t}, z^{w}_{t+1})-a_{t})^2}{2}}
\end{equation}
The forward dynamics objective function Equation~\ref{eq4} is defined as negative cosine similarity $\Delta$ between the predicted next latent state and the actual next latent state that encodes the next observation.
\begin{equation}\label{eq4}
    J({F}) = {\frac{\Delta(\tilde{z}^{s}_{t+1}, z^{s}_{t+1}) + \Delta(\bar{z}^{w}_{t+1}, z^{w}_{t+1})}{2}}
\end{equation}
\begin{table*}[t]
\centering
\resizebox{0.82\textwidth}{!}{
\begin{tabular}{cccccccc}
    \toprule
    & SAC & DrQ & CURL & SODA & PAD & SPD (Ours) \\
    \midrule
    \multicolumn{6}{l}{\textit{Data Efficiency (training and testing on Simple Distractor)}}\\
    \midrule
    Cheetah Run & 230.2$\pm$20.4 & 272.8$\pm$31.4 & \textbf{335.5$\pm$0.3} & 304.2$\pm$23.7 & 301.0$\pm$32.9 &  333.8$\pm$2.5 \\
    Finger Spin & 399.7$\pm$25.3 & 665.1$\pm$27.4 & 656.2$\pm$47.8 & 735.9$\pm$33.7 & 689.7$\pm$27.7 & \textbf{983.9$\pm$0.7} \\
    Hopper Hop & 92.4$\pm$5.6 & 91.5$\pm$31.8 & 73.6$\pm$27.5 & 86.4$\pm$43.5 & 125.3$\pm$86.5 & \textbf{152.5$\pm$6.0} \\
    Reacher Easy & 107.3$\pm$0.4 & 230.2$\pm$47.4 & 409.2$\pm$45.0 & 286.4$\pm$50.0 & 286.7$\pm$160.4 & \textbf{645.5$\pm$107.1} \\
    Walker Walk & 37.1$\pm$4.5 & 493.5$\pm$105.2 & \textbf{917.4$\pm$12.0} & 869.1$\pm$12.0 & 861.7$\pm$1.8 & 895.0$\pm$7.3 \\
    \midrule
    \multicolumn{6}{l}{\textit{Data Efficiency (training and testing on Natural Video)}}\\
    \midrule
    Cheetah Run & 136.4$\pm$22.4 & 63.8$\pm$19.7 & 118.2$\pm$38.2 & 74.0$\pm$31.0 & 171.0$\pm$113.8 &  \textbf{330.2$\pm$25.5} \\
    Finger Spin & 288.8$\pm$11.9 & 205.0$\pm$144.5 & 227.4$\pm$146.9 & 58.7$\pm$40.0 & 3.3$\pm$1.7 & \textbf{983.2$\pm$1.2} \\
    Hopper Hop & 33.1$\pm$7.1 & 0.0$\pm$0.0 & 9.7$\pm$5.4 & 0.3$\pm$0.2 & 0.7$\pm$0.6 & \textbf{164.3$\pm$14.1} \\
    Reacher Easy & 100.1$\pm$1.9 & 89.7$\pm$14.6 & 413.9$\pm$106.7 & 80.4$\pm$4.8 & 104.0$\pm$20.5 & \textbf{574.4$\pm$61.9} \\
    Walker Walk & 32.7$\pm$2.4 & 104.4$\pm$43.3 & 811.9$\pm$52.1 & 404.1$\pm$47.3 & 72.5$\pm$7.7 & \textbf{895.8$\pm$17.9} \\
    \midrule
    \multicolumn{6}{l}{\textit{Generalization (training on Simple Distractor but testing on Natural Video)}}\\
    \midrule
    Cheetah Run & 51.0$\pm$18.4 & 218.6$\pm$25.2 & 189.5$\pm$31.1 & 228.7$\pm$17.8 & 298.3$\pm$28.7 & \textbf{328.7$\pm$6.2} \\
    Finger Spin & 125.2$\pm$27.0 & 661.3$\pm$26.8 & 647.2$\pm$44.4 & 652.5$\pm$38.5 & 690.0$\pm$27.7 & \textbf{893.2$\pm$29.5} \\
    Hopper Hop & 14.8$\pm$5.2 & 81.4$\pm$30.0 & 42.7$\pm$23.6 & 59.2$\pm$32.0 & 112.7$\pm$67.7 & \textbf{134.6$\pm$6.2} \\
    Reacher Easy & 109.4$\pm$3.6 & 158.3$\pm$20.2 & 286.9$\pm$46.7 & 160.4$\pm$28.0 & 273.7$\pm$158.0 & \textbf{431.5$\pm$118.8} \\
    Walker Walk & 57.8$\pm$16.9 & 270.5$\pm$81.6 & 407.6$\pm$35.0 & 754.2$\pm$25.4 & 835.3$\pm$1.4 & \textbf{854.6$\pm$16.3} \\
    \bottomrule
\end{tabular}
}

\caption{Performance of SPD and baselines on five tasks in the DeepMind Control suite. We train for 500K environment steps on \textit{Simple Distractor} and \textit{Natural Video}. We evaluate the trained model on the same \textit{Simple Distractor} and \textit{Natural Video} for data efficiency experiments, and evaluate the model which is trained on \textit{Simple Distractor} on unseen \textit{Natural Video} for generalization experiments. The results show the mean and standard deviation over three different seeds.}

\label{table_dmc_result}
\end{table*}
The dynamics chaining objective function Equation~\ref{eq5} is defined as a combination of inverse dynamics and forward dynamics.
\begin{equation}\label{eq5}
    J({\psi}) = J({I}) + J({F})
\end{equation}
The Self-Predictive Dynamics (SPD) objective function is defined as a combination of dynamics chaining and discriminator as shown in Equation~\ref{eq6}, and it can send a training signal to the encoder $\phi$ to efficiently represent task-relevant features. 
\begin{equation}\label{eq6}
    J({\psi, \phi,D}) = \lambda_{\psi} J({\psi}) + \lambda_{A} J({\phi,D})
\end{equation}
where $\lambda_{\psi}$ and $\lambda_{A}$ are hyper parameters.\footnote{We have shown that SPD has good performance in a wide range of hyper parameter choices in the supplementary material.}

Algorithm~\ref{alg:algorithm} describes how SPD works. In the algorithm, $s_t, s_{t+1}$ are the image observations obtained by interacting with the environment.
We divide the training phase of SPD into two steps. First, train an encoder by optimizing SPD objective, and then train the RL policy.
We repeat this learning process and SPD objective functions refer to Equation~\ref{eq6}.
This algorithm version is based on an off-policy RL algorithm, such as Soft Actor-Critic (SAC)~\cite{haarnoja2018soft}, but our method (SPD) can work with any RL algorithms, as shown in the supplementary material. (such as on-policy algorithms like PPO~\cite{schulman2017proximal} and other off-policy algorithms like TD3~\cite{fujimoto2018addressing}).
\section{Experiments}
This section demonstrates how efficiently SPD can learn vision-based control tasks with distracting elements (task-irrelevant information) and can generalize well against unseen test environments.
On a set of continuous control tasks in the DeepMind Control suite, SPD shows excellent performance in most settings.
For CARLA~\cite{dosovitskiy2017carla}, a more realistic and autonomous driving environment with various distractors (e.g., shadows, changing weather, and light), we also show better performance than prior studies.
We benchmark SPD against the following algorithms; SAC is plain Soft Actor-Critic with no augmentation. DrQ~\cite{kostrikov2020image} applies data augmentations and regularized Q-function in SAC. CURL~\cite{srinivas2020curl} introduces a method of combining contrastive representation learning and RL.
SODA~\cite{hansen2021generalization} learns representation by maximizing the mutual information between augmented and non-augmented data. PAD~\cite{hansen2020self} fine-tunes representations at testing environments through self-supervision.
\begin{figure}[t]
\centering
\includegraphics[width=0.75\columnwidth]{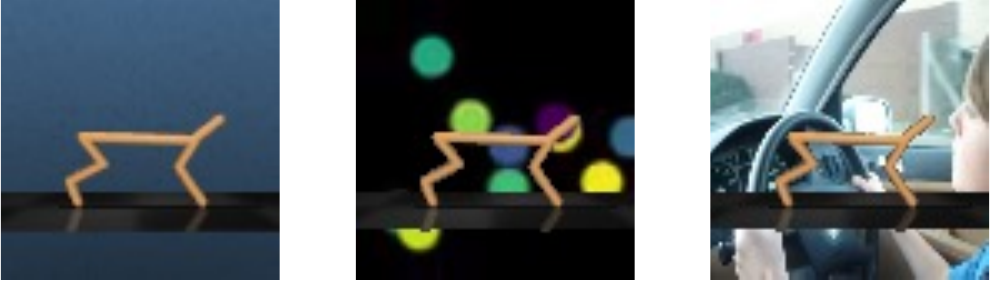}
\caption{We use three different background types. There are examples on a Cheetah task in the Deepmind Control suite; \textit{Default} (left), \textit{Simple Distractor} (center), and \textit{Natural Video} (right)}
\label{fig:dmc}
\end{figure}

\begin{figure*}[t]
\centering
\includegraphics[width=0.80\textwidth]{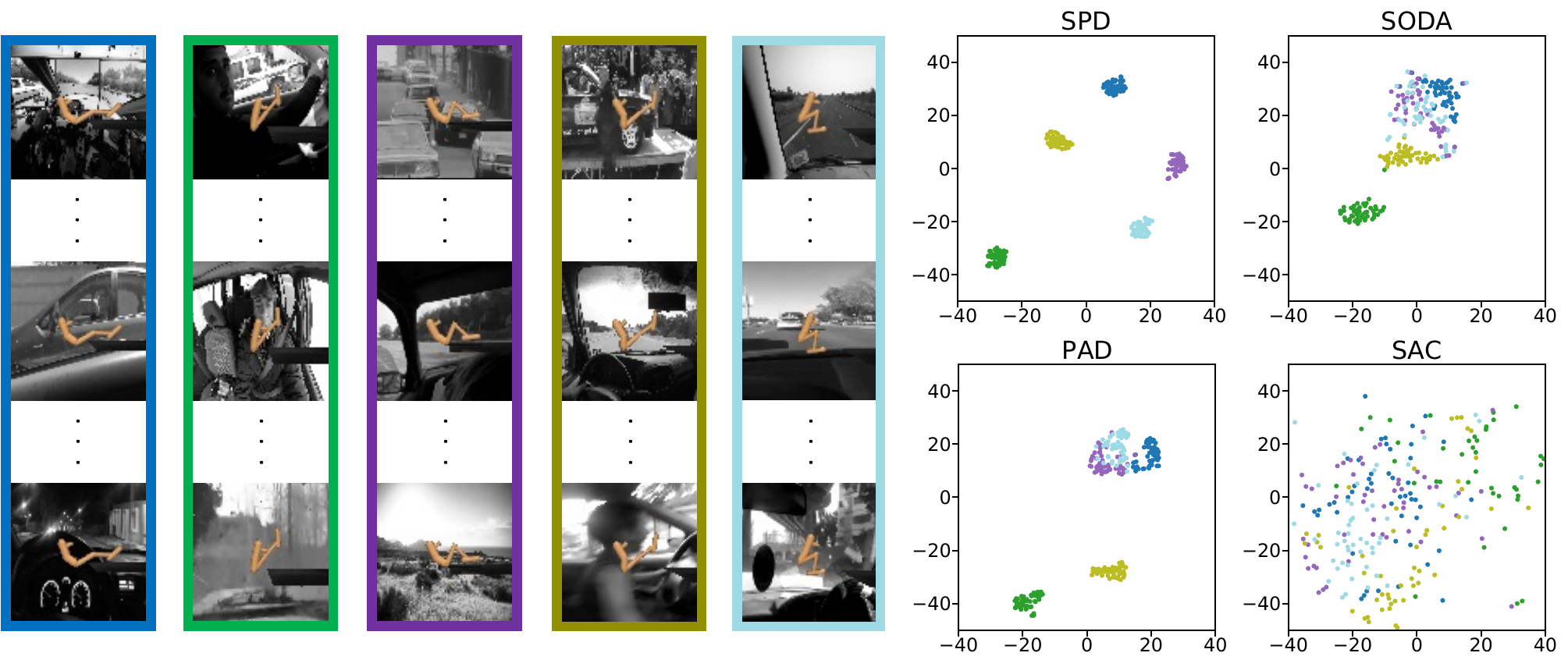}
\vspace{-0.2cm}
\caption{t-SNE of representations learned by SPD, SODA, PAD and SAC. Even if the background is dramatically different, SPD can encode behaviorally-equivalent observations (blue, green, violet, olive, sky blue) to be most closely located.}
\label{fig:tsne}
\end{figure*}
\subsection{Network Architecture}
\label{network_architecture}
We implement our SPD on top of Soft Actor Critic (SAC) for the visual input version~\cite{yarats2019improving} architecture, which updates the encoder only with Q-function back-propagation.
The RL parts Actor, Critic and the self-supervised part SPD share the Encoder $\phi$ which consists of 4 convolutional layers and 1 fully connected layer. Both Actor and Critic consists of 3 fully connected layers.
Dynamics chaining $\psi$ and Discriminator $D$ consists of 4 fully connected layers and 2 fully connected layers, respectively.
For CARLA, we modify the Encoder $\phi$ slightly. Implementation details and hyper parameters are in the supplementary material.

\subsection{DeepMind Control Suite}
The DeepMind Control suite is a vision-based simulator that provides a set of continuous control tasks. We experiment with nine tasks; Cheetah Run, Finger Spin, Hopper Hop, Reacher Easy, Walker Walk and additional tasks in the supplementary material. And we evaluate the performances on two metrics; one is Data efficiency and the other is Generalization. 
Each RL method is trained for 500K environment steps, and every 5,000 steps, we evaluated the currently trained model by calculating the average return for 10 episodes.
We trained each RL method over three different seeds. As shown in Table~\ref{table_dmc_result}, SPD shows performance similar to the best performance of the prior works in data efficiency experiments with lower distractions, but significantly outperforms the prior works in data efficiency experiments with higher distractions and generalization experiments. More experiment details are in the supplementary material.

\subsubsection{Data Efficiency}
For the data efficiency evaluation, we used two background configurations; \textit{Simple Distractor} and \textit{Natural Video}, as shown in Figure~\ref{fig:dmc}.
\textit{Simple Distractor} is a non-stationary background with randomly plotted circles with different colors. \textit{Natural Video} is also a non-stationary background which consists of real car-driving scenes in Kinetics dataset~\cite{kay2017kinetics}. In this evaluation, the test is carried out in the same environment (the same background setup) used for training. Basically, the higher the level of distraction, the lower the task performance.
As shown in Table~\ref{table_dmc_result}, SPD outperforms other baselines on \textbf{3} out of \textbf{5} tasks in the \textit{Simple Distractor} background, but \textbf{5} out of \textbf{5} tasks in the \textit{Natural Video} background.
For example, SPD achieves performance gains of \textbf{22\%} and \textbf{396\%} on the Hopper Hop, compared to the best performance among the other RL methods for each background setup.
The learning curves for task environments and the additional backgrounds are in the supplementary material.

\subsubsection{Generalization}
\label{Generalization}
In this experiment, we first trained each RL method in the \textit{Simple Distractor} background and then evaluated it in the \textit{Natural Video} background, which was not seen during the training phase.
The bottom row in Table~\ref{table_dmc_result} presents that the generalization performance for unseen observations.
SPD significantly outperforms other baselines for all environments. On Finger Spin, SPD achieves \textbf{29\%} higher performance than PAD which is fine-tuned for testing observations.
All nine environments results and their learning curves are provided in the supplementary material.
In Figure~\ref{fig:tsne}, we also visualize the state embedding of Hopper Hop using t-SNE. Even if unseen backgrounds are dramatically different, a well-generalized encoder should capture invariant features when observations are behaviorally equivalent. It has been shown that SPD can encode semantically similar observations to be most closely located.

\subsubsection{Ablation Studies}
\begin{figure}[t]
    \centering
    \includegraphics[width=0.95\columnwidth]{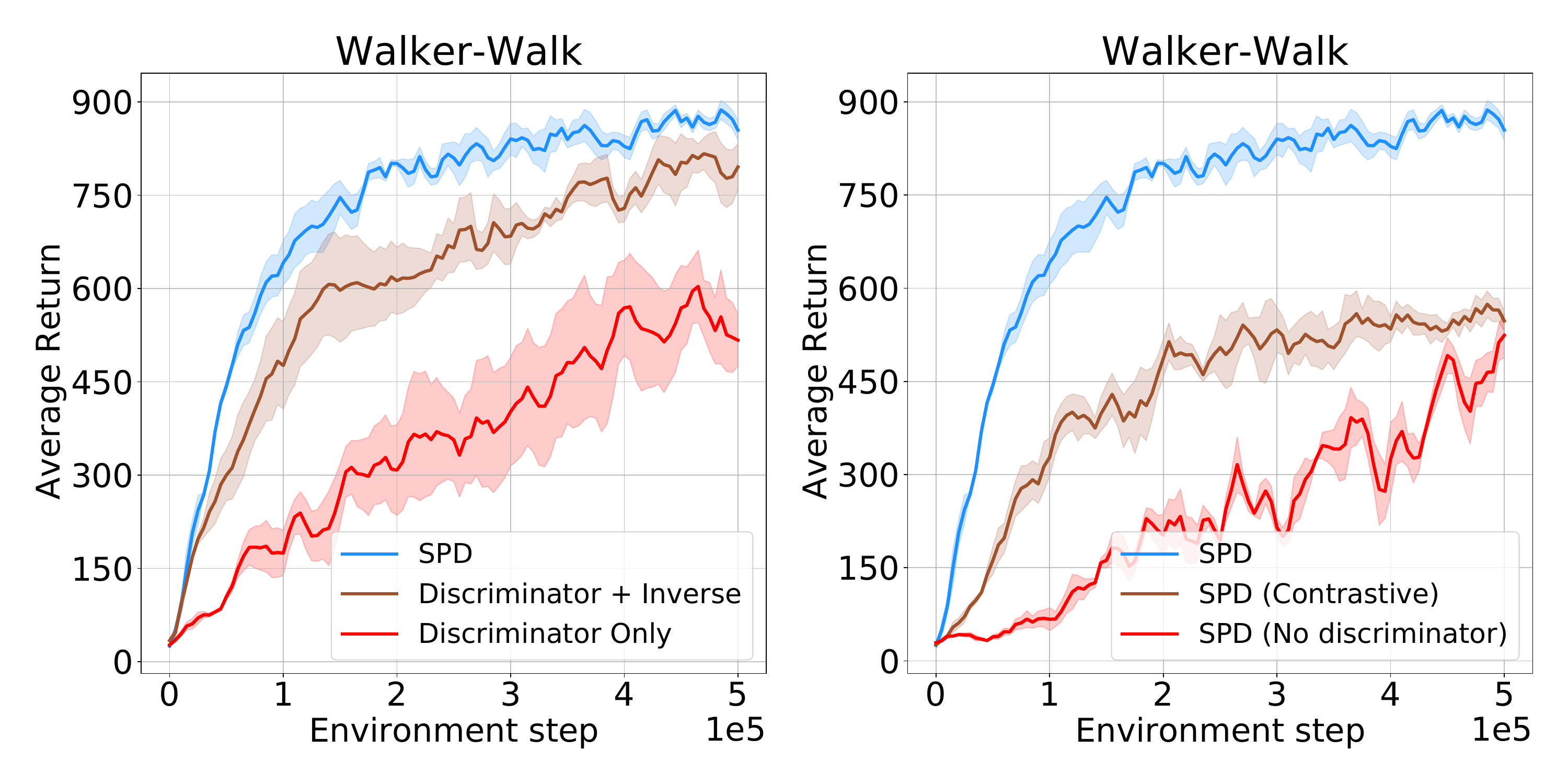}
    \vspace{-0.3cm}
\caption{(left) Ablation studies for SPD. We test the effects of discriminator, inverse dynamics and dynamics chaining. (right) Ablation studies for the different discriminator types. We show each ablation studies on three different seeds with 1.0 standard error shaded.}
\label{fig:ablation}
\end{figure}

We present the ablation studies to examine the synergy of our two-way data augmentations, discriminator, and dynamics chaining. Our ablation experiment is conducted in the same environment setup as the Generalization experiment.
In Figure~\ref{fig:ablation} (left), Discriminator Only stands for SAC with two-way data augmentations and the discriminator but no dynamics chaining. Discriminator + Inverse consists of two-way data augmentations, the discriminator, and the inverse dynamics (without the forward dynamics). The performance of Discriminator Only shows the lowest generalization performance. Discriminator + Inverse shows the performance can be highly improved because of the inverse dynamics. Although the role of the inverse dynamics greatly affects the performance, there is no doubt that our full integration (SPD) achieves the best performance.

In another ablation test, we try to analyze the role of the discriminator. We compare SPD to a version without the discriminator and a version with the contrastive learning method~\cite{srinivas2020curl}. Figure~\ref{fig:ablation} (right) shows clear differences in achieved task performance according to the different discriminator settings. Using the contrastive learning method improves performance compared to no-discriminator version. However, SPD using a relativistic GAN as the discriminator outperforms the version using the contrastive learning method much better.

\subsection{CARLA Environment}
\begin{figure}[t]
\centering
\includegraphics[width=0.40\textwidth]{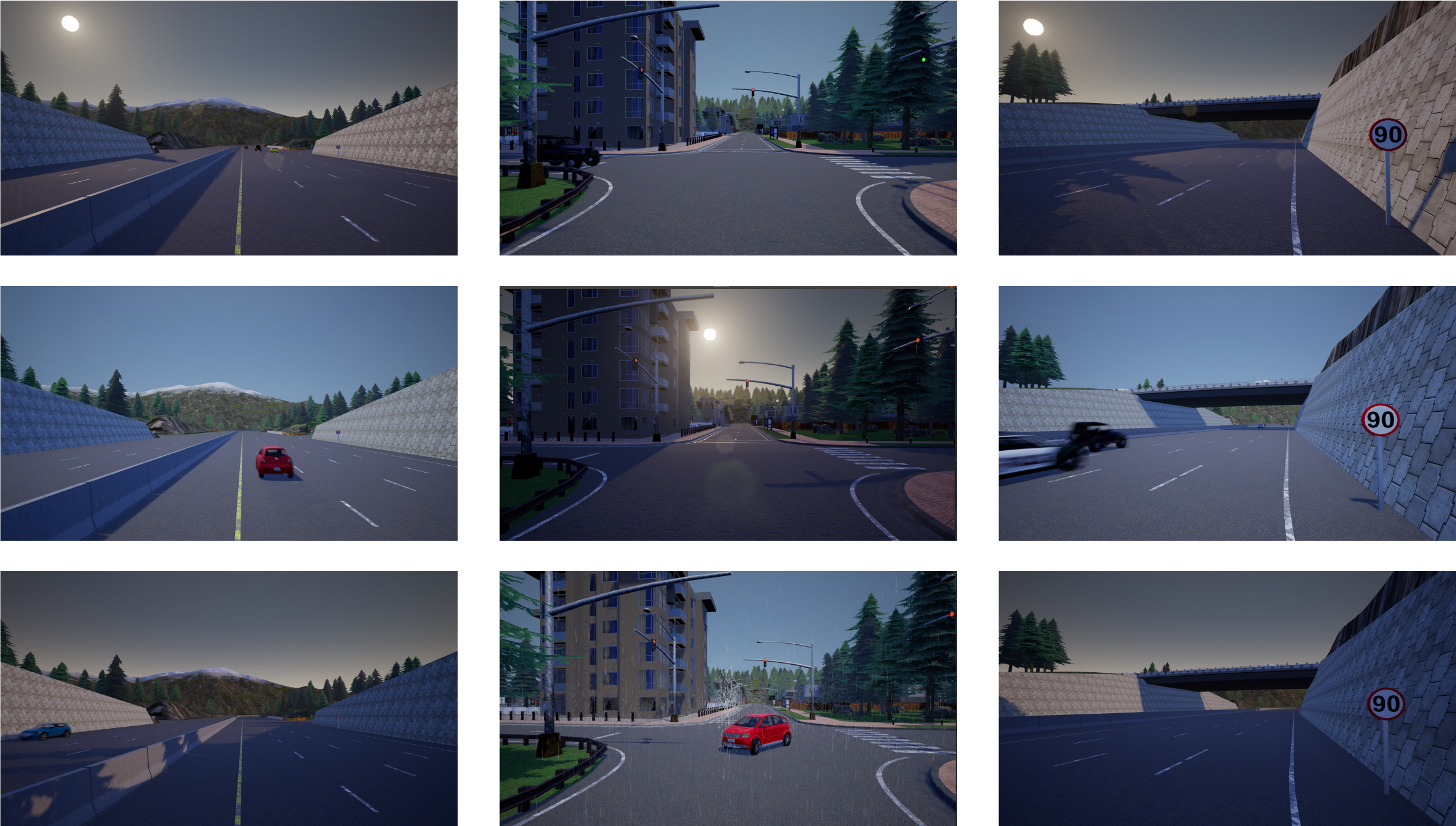}
\vspace{-0.05cm}
\caption{Scenes in CARLA simulations classified as Highway (left column), Town (center column) and Bridge (right column). Each column is captured in the same spot but contains different task-irrelevant information such as the Sun, rain, shadows, clouds, etc.}
\label{fig:CARLApic}
\end{figure}

\begin{figure}[t]
\centering
\includegraphics[width=0.43\textwidth]{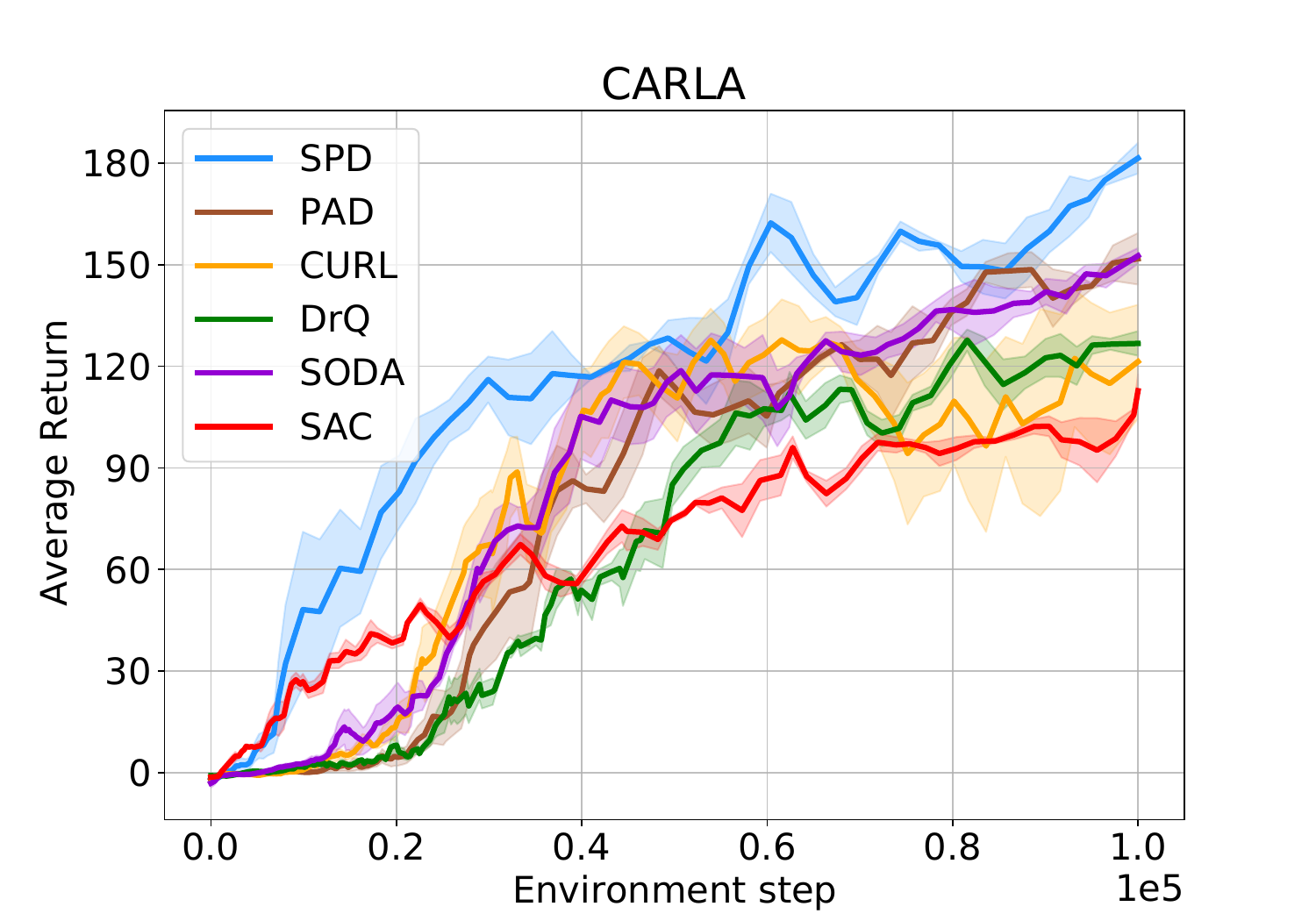}
\vspace{-0.2cm}
\caption{Performance comparison in the autonomous driving environment CARLA. SPD outperforms all other baselines.}
\label{fig:carlaresult}
\end{figure}

CARLA is a first-person view simulator for studying autonomous driving systems. In the CARLA simulations, we can evaluate the performance of RL methods on more realistic visual observations. As shown in Figure~\ref{fig:CARLApic}, there are diverse types of distractors (e.g., the Sun, rain, shadows, clouds, etc.) around the agent, and it changes dynamically with every episode, and even within the same episode. Therefore, it becomes more important to extract control-related features (e.g., road, collision, speed, brake, steer, etc.). The basic experimental setup is configured the same as DBC~\cite{zhang2020learning}.
Visual observation is a 300 degree view from the vehicle roof and the image size is 3$\times$84$\times$420. The reward is defined by the function of driving distance, speed, and the penalty of collision, steering and breaking.
Each method is trained for 100K environment steps, and the average return for 20 test episodes is calculated. We run each RL method across three seeds.
Figure~\ref{fig:carlaresult} shows the performance comparison with three seeds in CARLA.
SODA performs better than other baselines and is comparable with the performance of SPD, but SPD learns much faster and achieves the highest performance.
For another comparison of representation quality, we suggest the representation distance in latent space between two observations. We can intuitively assume that the representation distance should be close if their task-relevant context is similar regardless of other distracting elements. We first took 50 random observations at three locations; Highway, Town, and Bridge in CARLA.
We repeatedly collected observations from almost the same spots, but these observation characteristics change because of varying task-irrelevant information (e.g., the Sun, shadows, clouds, rain, car types \& colors, etc.), as shown in Figure~\ref{fig:CARLApic}.
We measured the L2 distance in the latent space between various observations obtained under behaviorally identical circumstances.
Table~\ref{table1} presents the average representation distance normalized to the SPD result. It shows that SPD has minimal average distance compared to other studies, and we believe this is why our method performs best.

\begin{table}[t]
\centering
\resizebox{.45\textwidth}{!}{
\begin{tabular}{c|cccccc}
    \toprule
    & SAC & DrQ & CURL & SODA & PAD & SPD \\
    \midrule
    Highway & 3.86 & 2.20 & 2.38 & 1.74 & 1.83 &\textbf{1.00} \\
    Town & 6.23 & 3.57 & 3.86 & 2.43 & 2.47 &\textbf{1.00}\\
    Bridge & 3.82 & 1.61 & 1.57 & 1.25 & 1.41 &\textbf{1.00} \\
    \bottomrule
\end{tabular}}
\vspace{-0.1cm}
\caption{Average representation distance of latent space according to task-irrelevant information changes in CARLA simulations. (The numbers are normalized to SPD)}
\label{table1}
\end{table}
\section{Conclusion}
In this work, we propose a novel representation learning method for vision-based RL. Our proposed Self-Predictive Dynamics based on two-way (weak and strong) data augmentations can significantly improve the data efficiency and generalization performance when operating on highly complex or unseen observations. In the future, we plan to design a sequence-based generalization approach such as representing a series of image inputs and predicting multi-step dynamics chaining in latent space.

\section*{Acknowledgements}
This work was supported partly by the Institute of Information and Communications Technology Planning and Evaluation (IITP) grant funded by the Korea Government (MSIT) (No. 2022-0-01045, Self-directed Multi-Modal Intelligence for solving unknown, open domain problems), (No. 2022-0-00688, AI Platform to Fully Adapt and Reflect Privacy-Policy Changes), (No. 2020-0-00973, Reconstruction of Non-Line-of-Sight Scene for VR/AR Contents) and (No. 2019-0-00421, Artificial Intelligence Graduate School Program(Sungkyunkwan University)).

\bibliographystyle{named}
\bibliography{ijcai22}
\clearpage
\appendix
\begin{center}
\textbf{\huge Supplementary Material}\\
\end{center}

\section{Data Augmentation ablations}
\begin{figure}[!ht]
\centering
\includegraphics[width=0.9\columnwidth]{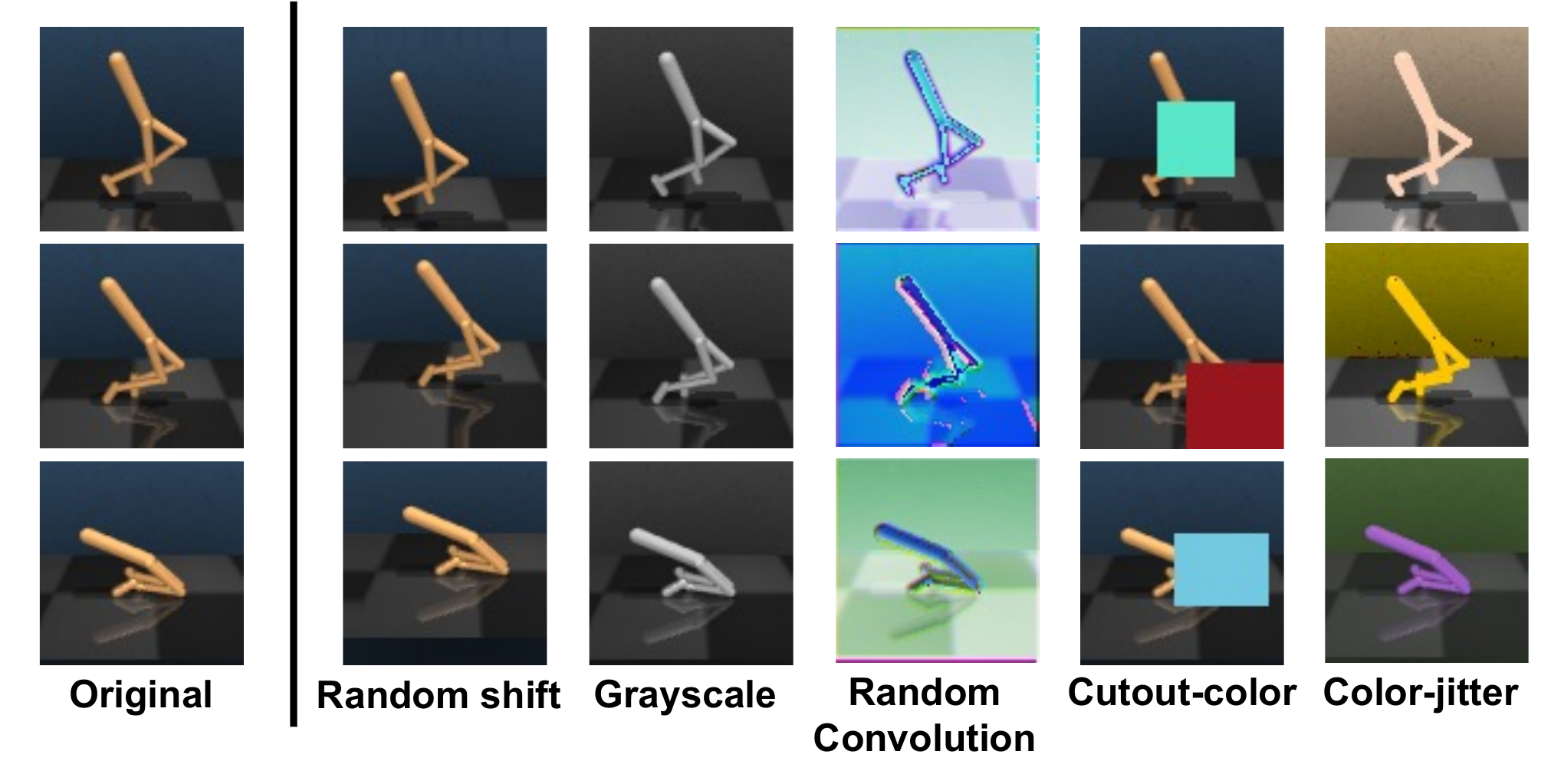}
\vspace{-0.2cm}
\caption{Data augmentations used in our framework: (From the left) Random-shift, Grayscale, Random Convolution, Cutout-color, and Color-jitter}
\label{fig:image_aug}
\end{figure}

\begin{figure}[!ht]
    \centering
    \includegraphics[width=0.9\columnwidth]{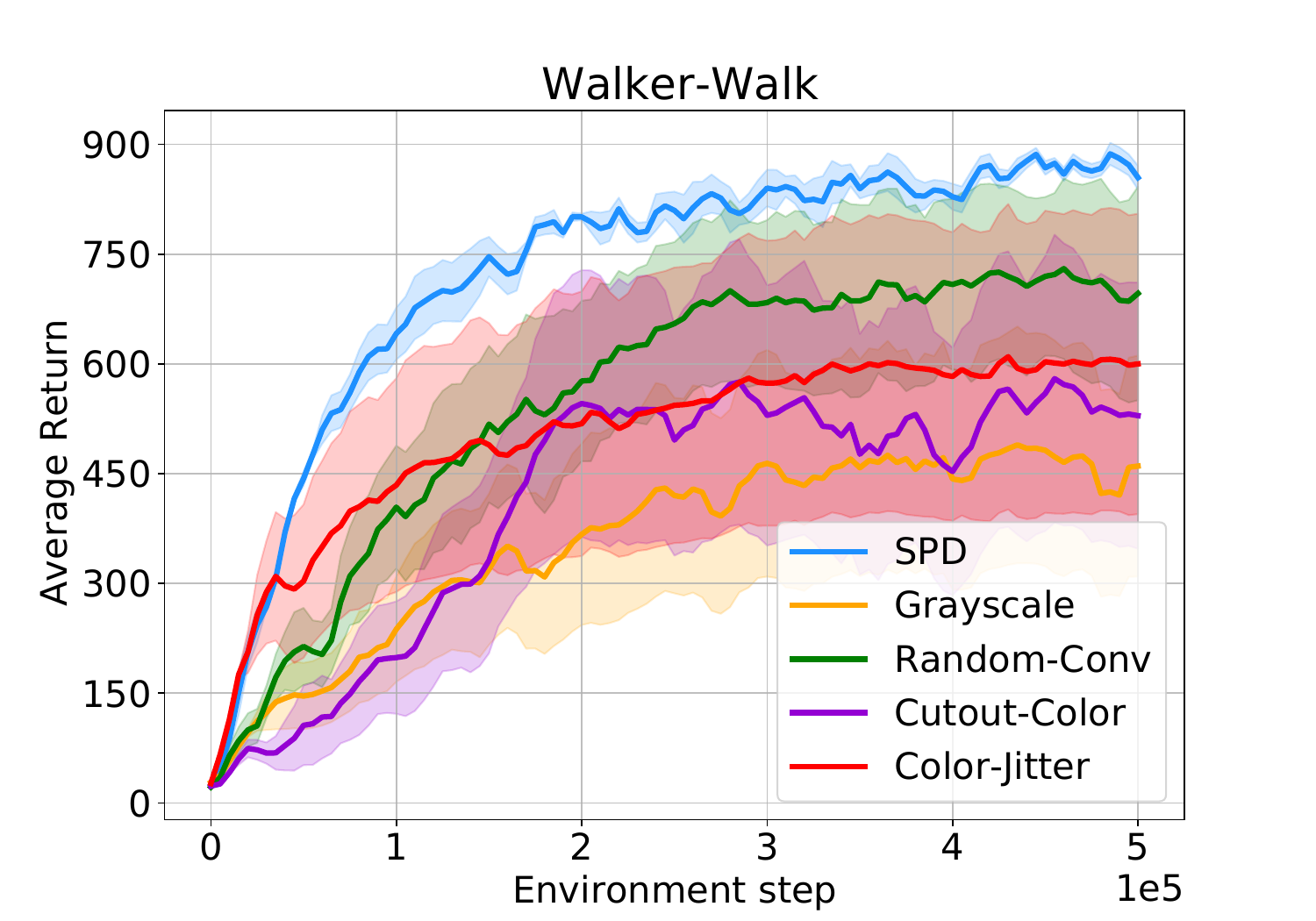}
\caption{Using multiple strong augmentation techniques together shows better performance than using a single strong augmentation. We show the learning curves of each experiments on three different seeds with 1.0 standard error shaded.}
\label{fig:augmentation}
\end{figure}
For our strong augmentation, SPD uses random-shift and randomly adds one of grayscale, random-convolution, cutout-color and color-jitter every mini-batch.
Figure~\ref{fig:augmentation} compares the performance when using only one type of strong augmentation each.
Random-Convolution seems to have the greatest impact on performance, but when all the data augmentations are used together shows the best performance.

\section{Apply Self-Predictive Dynamics (SPD) to Other RL Algorithms}\label{appendix_other}
\begin{figure}[t]
    \centering
    \includegraphics[width=0.9\columnwidth]{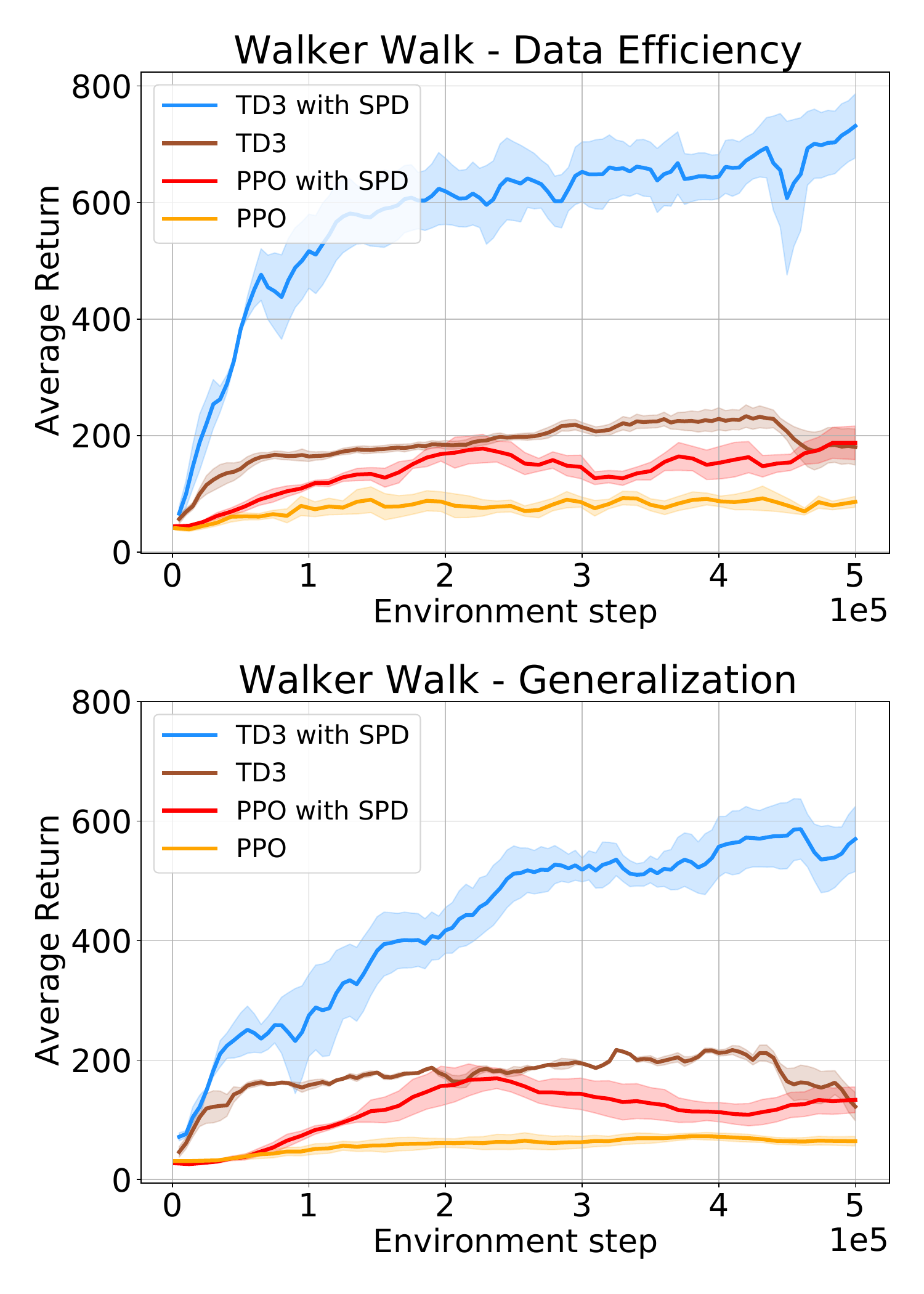}
\vspace{-0.2cm}
\caption{(top) Data Efficiency evaluation, (bottom) Generalization evaluation on the DeepMind Control suite (Walker Walk). TD3 with SPD (blue), standard TD3 (brown), PPO with SPD (red), standard PPO (orange). We show the learning curves of each experiments on three different seeds with 1.0 standard error shaded.}
\label{fig:td3_ppo}
\end{figure}

In this section, we show whether SPD improves the data efficiency and generalization performance using other RL algorithms. We have replaced SAC with a different off-policy algorithms, TD3~\cite{fujimoto2018addressing}, and one of the on-policy algorithms, PPO~\cite{schulman2017proximal}.
As shown in Figure~\ref{fig:td3_ppo}, our framework is helpful to improve Data Efficiency and Generalization performance no matter what RL algorithms we apply.

\section{All of DeepMind Control suite results}
\label{additional_dmc}
\begin{figure*}[!htp]
\begin{center}
\includegraphics[width=0.80\textwidth]{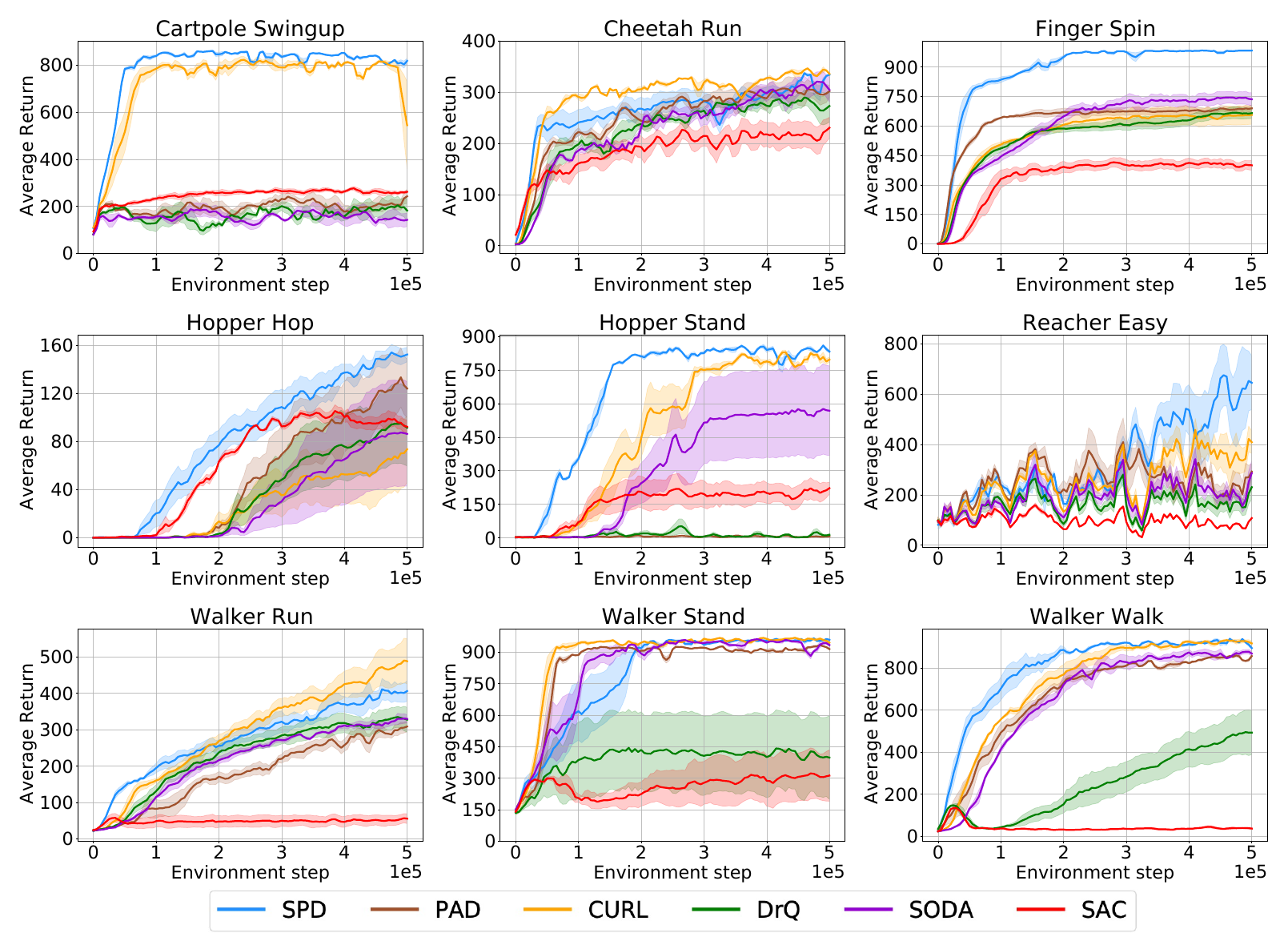}
\end{center}
\vspace{-0.4cm}
\caption{Results of data efficiency evaluation for SPD and baselines on \textit{Simple Distractor} background. We show the learning curves of each tasks on three different seeds with 1.0 standard error shaded.}
\label{fig:app_i}
\end{figure*}

\begin{figure*}[!htp]
\begin{center}
\includegraphics[width=0.80\textwidth]{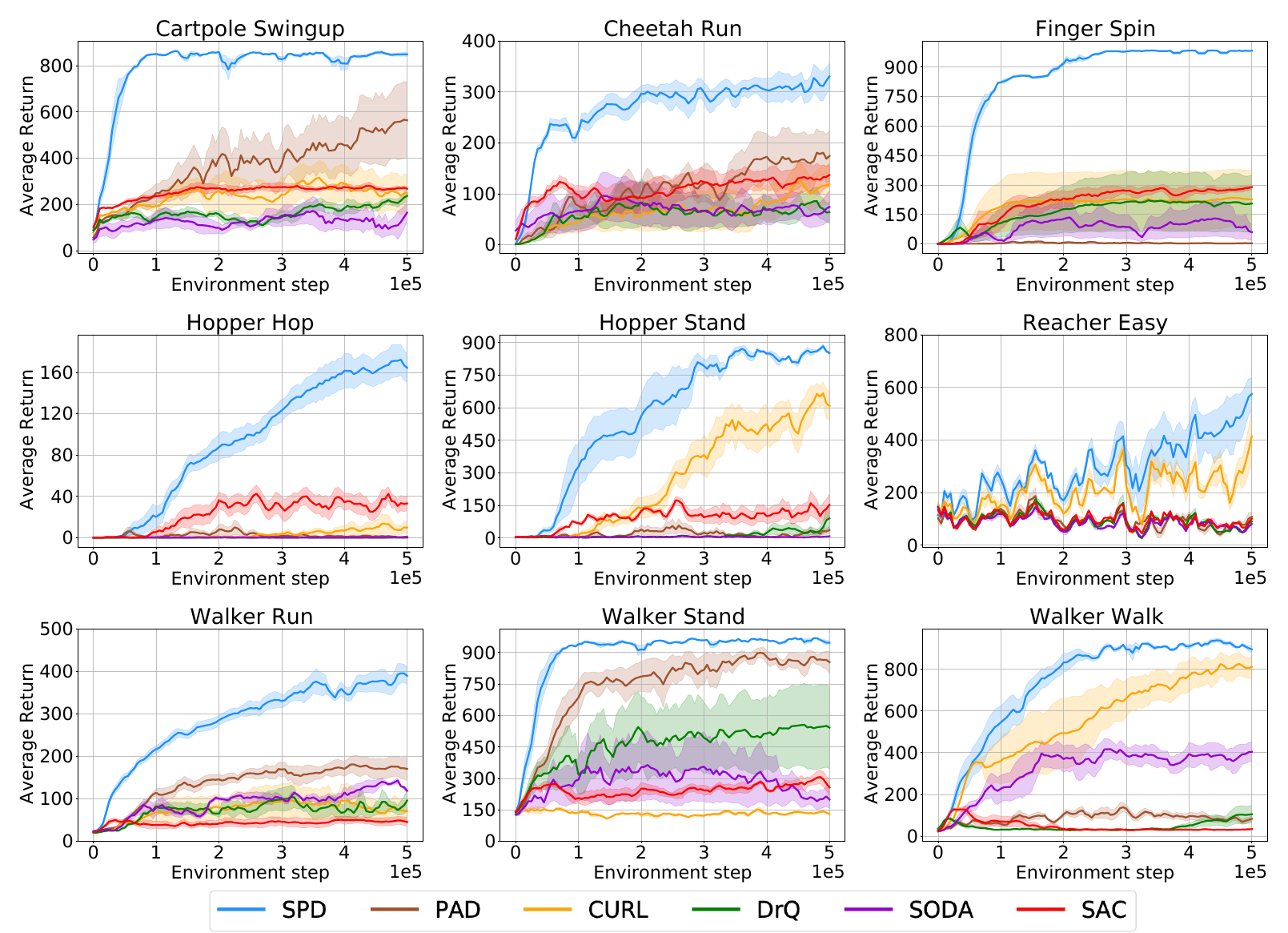}
\end{center}
\vspace{-0.4cm}
\caption{Results of data efficiency evaluation for SPD and baselines on \textit{Natural Video} background. We show the learning curves of each tasks on three different seeds with 1.0 standard error shaded.}
\label{fig:app_n}
\end{figure*}

\begin{figure*}[!htp]
\begin{center}
\includegraphics[width=0.80\textwidth]{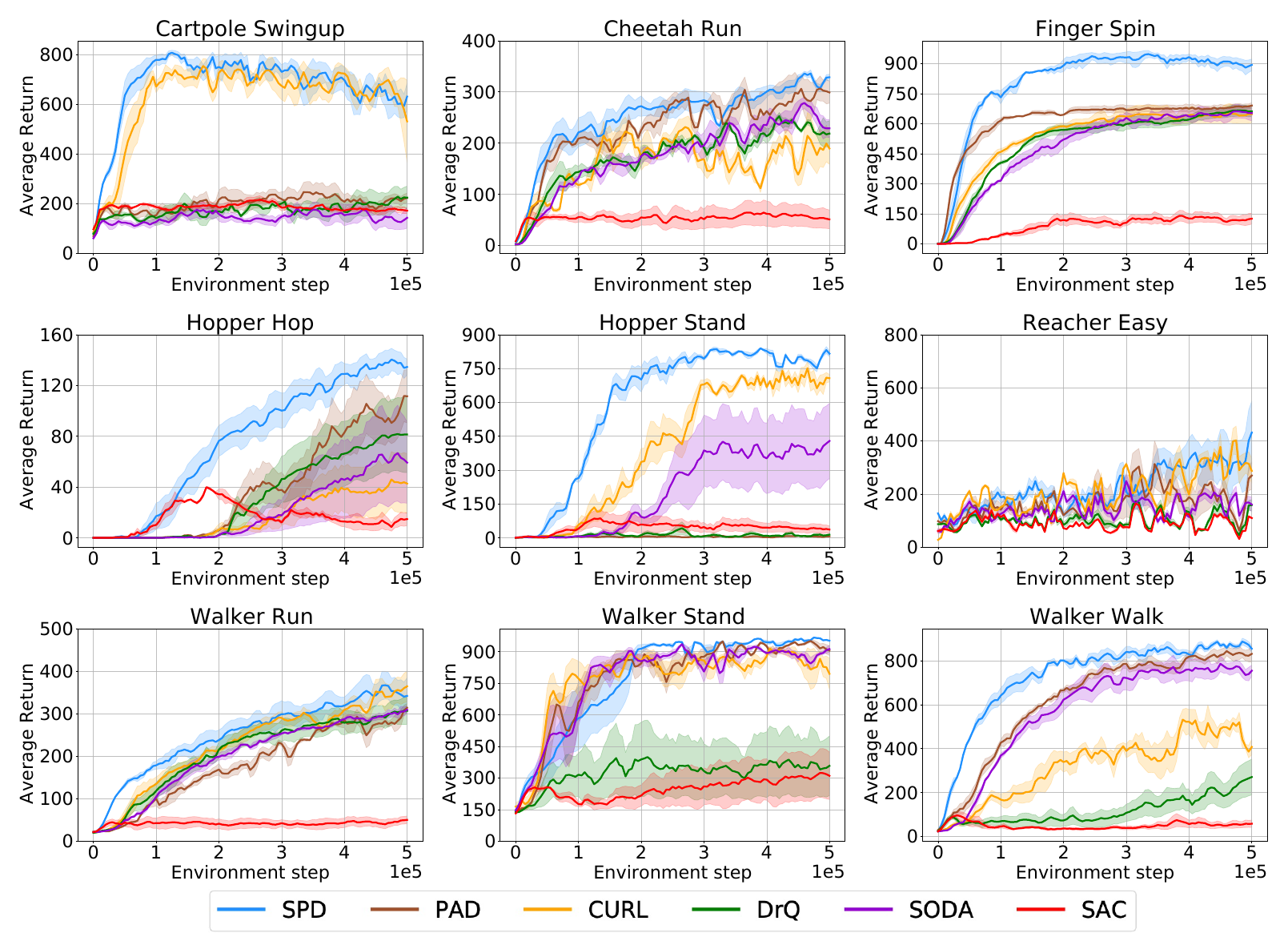}
\end{center}
\vspace{-0.4cm}
\caption{Results of generalization evaluation for SPD and baselines. We show the learning curves of each tasks on three different seeds with 1.0 standard error shaded.}
\label{fig:app_g}
\end{figure*}

Table~\ref{additional_table_dmc_result} gives comprehensive performance of SPD and baselines for all nine environments.
Figure~\ref{fig:app_i}, \ref{fig:app_n} show learning curves of Data efficiency on \textit{Simple Distractor} and \textit{Natural Video} background setting for nine environments.
Figure~\ref{fig:app_g} shows learning curves of Generalization for nine environments.
We evaluate performance of Generalization through average return differences when evaluating on \textit{Simple Distractor} and \textit{Natural Video} background after training on \textit{Simple Distractor} background.
We note that the original CURL, SODA and PAD use Random Crop which randomly crops an 84 x 84 image from a 100 x 100 simulation-rendered image during training. However, these methods use a center crop of an 84 x 84 image from a 100 x 100 image for evaluation. In the DeepMind Control suit environments, robots are always located in the center of the simulation images. Therefore, the center crop easily removes background areas. To evaluate the performance on distracting backgrounds and unseen backgrounds, such a center crop is NOT fair. In our experiments, we replace Rrandom Crop (for training) \& Center Crop (for evaluation) with Random-shift (for training) \& No augmentation (for evaluation).

\begin{table*}[!htp]
\centering
\begin{tabular}{cccccccc}
    \toprule
    & SAC & DrQ & CURL & SODA & PAD & SPD (Ours) \\
    \midrule
    \multicolumn{6}{l}{\textit{Data Efficiency (training and testing on Simple Distractor)}}\\
    \midrule
    Cartpole Swingup & 261.7$\pm$6.6 & 182.1$\pm$48.5 & 544.4$\pm$185.2 & 142.6$\pm$27.8 & 242.3$\pm$20.2 & \textbf{817.7$\pm$10.7} \\
    Cheetah Run & 230.2$\pm$20.4 & 272.8$\pm$31.4 & \textbf{335.5$\pm$0.3} & 304.2$\pm$23.7 & 301.0$\pm$32.9 &  333.8$\pm$2.5 \\
    Finger Spin & 399.7$\pm$25.3 & 665.1$\pm$27.4 & 656.2$\pm$47.8 & 735.9$\pm$33.7 & 689.7$\pm$27.7 & \textbf{983.9$\pm$0.7} \\
    Hopper Hop & 92.4$\pm$5.6 & 91.5$\pm$31.8 & 73.6$\pm$27.5 & 86.4$\pm$43.5 & 125.3$\pm$86.5 & \textbf{152.5$\pm$6.0} \\
    Hopper Stand & 221.8$\pm$27.0 & 13.9$\pm$4.7 & 796.7$\pm$11.3 & 568.1$\pm$200.3 & 6.3$\pm$1.2 & \textbf{832.1$\pm$6.4} \\
    Reacher Easy & 107.3$\pm$0.4 & 230.2$\pm$47.4 & 409.2$\pm$45.0 & 286.4$\pm$50.0 & 286.7$\pm$160.4 & \textbf{645.5$\pm$107.1} \\
    Walker Run & 55.4$\pm$13.5 & 327.3$\pm$34.1 & \textbf{488.1$\pm$62.0} & 329.6$\pm$11.3 & 305.0$\pm$14.0 & 406.1$\pm$28.1 \\
    Walker Stand & 312.6$\pm$122.1 & 398.2$\pm$198.8 & 944.9$\pm$8.2 & 933.8$\pm$8.9 & 911.7$\pm$2.9 & \textbf{958.6$\pm$2.6} \\
    Walker Walk & 37.1$\pm$4.5 & 493.5$\pm$105.2 & \textbf{917.4$\pm$12.0} & 869.1$\pm$12.0 & 861.7$\pm$1.8 & 895.0$\pm$7.3 \\
    \midrule
    \multicolumn{6}{l}{\textit{Data Efficiency (training and testing on Natural Video)}}\\
    \midrule
    Cartpole Swingup & 267.5$\pm$8.7 & 236.3$\pm$25.1 & 246.9$\pm$80.0 & 163.9$\pm$40.7 & 558.3$\pm$344.9 & \textbf{847.8$\pm$9.3} \\
    Cheetah Run & 136.4$\pm$22.4 & 63.8$\pm$19.7 & 118.2$\pm$38.2 & 74.0$\pm$31.0 & 171.0$\pm$113.8 &  \textbf{330.2$\pm$25.5} \\
    Finger Spin & 288.8$\pm$11.9 & 205.0$\pm$144.5 & 227.4$\pm$146.9 & 58.7$\pm$40.0 & 3.3$\pm$1.7 & \textbf{983.2$\pm$1.2} \\
    Hopper Hop & 33.1$\pm$7.1 & 0.0$\pm$0.0 & 9.7$\pm$5.4 & 0.3$\pm$0.2 & 0.7$\pm$0.6 & \textbf{164.3$\pm$14.1} \\
    Hopper Stand & 152.1$\pm$49.9 & 90.4$\pm$56.9 & 608.3$\pm$67.5 & 7.3$\pm$24.0 & 25.5$\pm$1.2 & \textbf{851.5$\pm$5.2} \\
    Reacher Easy & 100.1$\pm$1.9 & 89.7$\pm$14.6 & 413.9$\pm$106.7 & 80.4$\pm$4.8 & 104.0$\pm$20.5 & \textbf{574.4$\pm$61.9} \\
    Walker Run & 45.0$\pm$10.7 & 95.7$\pm$24.1 & 70.2$\pm$24.9 & 118.6$\pm$5.3 & 170.0$\pm$54.3 & \textbf{389.1$\pm$19.4} \\
    Walker Stand & 256.3$\pm$40.9 & 541.9$\pm$196.2 & 128.8$\pm$1.3 & 198.8$\pm$41.6 & 848.5$\pm$109.4 & \textbf{946.3$\pm$10.7} \\
    Walker Walk & 32.7$\pm$2.4 & 104.4$\pm$43.3 & 811.9$\pm$52.1 & 404.1$\pm$47.3 & 72.5$\pm$7.7 & \textbf{895.8$\pm$17.9} \\
    \midrule
    \multicolumn{6}{l}{\textit{Generalization (training on Simple Distractor but testing on Natural Video)}}\\
    \midrule
    Cartpole Swingup & 172.2$\pm$11.2 & 224.8$\pm$46.6 & 531.9$\pm$165.2 & 143.6$\pm$41.9 & 223.0$\pm$24.2 & \textbf{632.0$\pm$54.5} \\
    Cheetah Run & 51.0$\pm$18.4 & 218.6$\pm$25.2 & 189.5$\pm$31.1 & 228.7$\pm$17.8 & 298.3$\pm$28.7 & \textbf{328.7$\pm$6.2} \\
    Finger Spin & 125.2$\pm$27.0 & 661.3$\pm$26.8 & 647.2$\pm$44.4 & 652.5$\pm$38.5 & 690.0$\pm$27.7 & \textbf{893.2$\pm$29.5} \\
    Hopper Hop & 14.8$\pm$5.2 & 81.4$\pm$30.0 & 42.7$\pm$23.6 & 59.2$\pm$32.0 & 112.7$\pm$67.7 & \textbf{134.6$\pm$6.2} \\
    Hopper Stand & 37.1$\pm$16.5 & 14.4$\pm$4.5 & 707.2$\pm$16.6 & 429.2$\pm$166.5 & 5.7$\pm$0.5 & \textbf{815.7$\pm$12.8} \\
    Reacher Easy & 109.4$\pm$3.6 & 158.3$\pm$20.2 & 286.9$\pm$46.7 & 160.4$\pm$28.0 & 273.7$\pm$158.0 & \textbf{431.5$\pm$118.8} \\
    Walker Run & 50.2$\pm$9.1 & 306.7$\pm$31.9 & \textbf{364.2$\pm$36.4} & 309.1$\pm$9.3 & 311.0$\pm$14.0 & 341.7$\pm$42.5 \\
    Walker Stand & 310.5$\pm$112.4 & 358.0$\pm$141.6 & 749.4$\pm$82.0 & 912.2$\pm$17.9 & 909.7$\pm$2.5 & \textbf{951.9$\pm$1.3} \\
    Walker Walk & 57.8$\pm$16.9 & 270.5$\pm$81.6 & 407.6$\pm$35.0 & 754.2$\pm$25.4 & 835.3$\pm$1.4 & \textbf{854.6$\pm$16.3} \\
    \bottomrule
\end{tabular}
\caption{Comprehensive performance of SPD and baselines on nine tasks in the DeepMind Control suite. The results show the mean and standard deviation over three different seeds.}
\vspace{-0.2cm}
\label{additional_table_dmc_result}
\end{table*}

\newpage
\section{Implementation Details}
\label{implementation_details}
In this section, we explain the implementation details for SPD in the DeepMind Control suite and the CARLA.
We use Soft Actor Critic (SAC)~\cite{haarnoja2018soft} which is modified by ~\cite{yarats2019improving} and same encoder architecture as in~\cite{zhang2020learning} for DeepMind Control suite and CARLA.
For details on the SAC, the reader is referred to \cite{haarnoja2018soft}.
We use Pytorch implementation of SAC~\cite{kostrikov2020image} and build SPD on top of it. 
We augment with a shared encoder between the actor, critic, encoder $\phi$, SPD (Inverse Dynamics $I$, Forward Dynamics $F$, Discriminator $D$).

\subsection{Network architecture details}
Our encoder consists of four CNN layers with 3$\times$3 kernels, 32 channels and set the stride to 1 for each layer, except set the first layer's stride to 2. And then we apply ReLU activation function to all CNN layers. Finally, output of CNN layers is fed into a fully-connected layer normalized by LayerNorm and apply tanh activation to output of fully-connected layer. This output is 50 dimensional latent vector.
The actor, critic, dynamics chaining and discriminator networks share encoder $\phi$.
The critic (Q-function) consists of three-layer MLP with applied ReLU to all layers except the last layer, and size of hidden layers is 256.
The actor also consists of MLP architecture similar to critic and the final output is mean and covariance for the diagonal Gaussian, which represent the policy.
Inverse dynamics $I$, forward dynamics $F$ and discriminator $D$ networks consist of two-layer MLP with applied ReLU to first layers and size of hidden layers is 256. Output of these networks apply tanh activation.
When training in a CARLA environment, we modify encoder slightly. This is the same as used on DeepMind Control suite, except that stride is set to 2 all CNN layers.

\subsection{Experimental setup}
First, our agent collects 1000 observations of size 3$\times$84$\times$84 in DeepMind Control suite and size 3$\times$420$\times$420 using a random policy. 
After 1000 seed steps, agent is updated for each true environment step (when an episode length is 1000 steps, if action repeat is 2, true environment step is 500).
All methods are trained for 500,000 steps (DeepMind Control suite) or 100,000 steps (CARLA).
During training, the average return for 10 episodes is calculated to evaluate every 5000 true environment steps (DeepMind Control suite) or the average return for 20 episodes is calculated to evaluate every 10 episodes (CARLA).

\subsection{Data augmentations}
In common, we use an image observation as an 3-stack of consecutive frames. And then we normalize it by dividing by 255.
Data augmentations described in Figure~\ref{fig:image_aug} is applied to the normalized image.
We apply augmentation to images sampled from the buffer or a recent trajectory only during training procedure, not environment interaction procedure.
In the DeepMind Control suite, when Random shift is applied, all sides are padded by 4 pixels, but in the CARLA environment, the top and bottom sides are padded by 4 pixels, and the left and right sides are padded by 20 pixels according to the image ratio.

\subsection{Hyper-parameters}
When applying SPD, we adopt hyper parameters used in \cite{yarats2019improving}.
We detail all hyper parameters used for DeepMind Control suite and CARLA environments in Table~\ref{table:hyper1} and Table~\ref{table:hyper2}.
\begin{table}[!ht]
\centering
\begin{tabular}{c|c}
    \toprule
    \textbf{hyper parameter} & \textbf{Value}\\
    \midrule
    Frame & 3$\times$84$\times$84 \\
    Seed steps & 1000 \\
    Stacked frames & 3\\
    Action repeat & 2(walker, finger) \\
    & 4(otherwise) \\
    & 8(cartpole) \\
    Batch size & 128 \\
    Replay buffer size & 100,000 \\
    Number of training steps & 500,000\\
    Discount factor & 0.99 \\
    Optimizer & Adam \\
    Episode length & 1000 \\
    Actor, Critic Learning rate & $10^{-3}$ \\
    Dynamics chaining learning rate & $10^{-3}$ \\
    Encoder learning rate & $10^{-3}$ \\
    Discriminator learning rate & $10^{-3}$ \\
    Critic target update frequency & 2 \\
    Critic Q-function soft-update rate & 0.01 \\
    Actor update frequency & 2 \\
    Actor log stddev bounds & [-10, 2] \\
    Init temperature & 0.1 \\
    $\lambda_{\psi}$ & 0.1 \\ 
    $\lambda_{A}$ & 0.001\\
    \bottomrule
\end{tabular}
\vspace{-0.2cm}
\caption{Hyper parameters used for the DeepMind Control suite experiments.}
\label{table:hyper1}
\end{table}

\newpage
\begin{table}[!ht]
\centering
\begin{tabular}{c|c}
    \toprule
    \textbf{hyper parameter} & \textbf{Value}\\
    \midrule
    Frame & 3$\times$84$\times$420 \\
    Seed steps & 100 \\
    Stacked frames & 3\\
    Action repeat & 4 \\
    Batch size & 128 \\
    Replay buffer size & 100,000 \\
    Number of training steps & 100,000\\
    Discount factor & 0.99 \\
    Optimizer & Adam \\
    Episode length & 1000 \\
    Actor, Critic Learning rate & $10^{-3}$ \\
    Dynamics chaining learning rate & $10^{-3}$ \\
    Encoder learning rate & $10^{-3}$ \\
    Discriminator learning rate & $10^{-3}$ \\
    Critic target update frequency & 2 \\
    Critic Q-function soft-update rate & 0.01 \\
    Actor update frequency & 2 \\
    Actor log stddev bounds & [-10, 2] \\
    Init temperature & 0.1 \\
    $\lambda_{\psi}$ & 0.1 \\ 
    $\lambda_{A}$ & 0.001\\
    \bottomrule
\end{tabular}

\vspace{-0.2cm}
\caption{Hyper parameters used for the CARLA experiments.}
\label{table:hyper2}
\label{table:hyperall}
\end{table}


\section{Hyper-parameter Sensitivity}
\begin{figure}[!ht]
    \centering
    \includegraphics[width=0.45\textwidth]{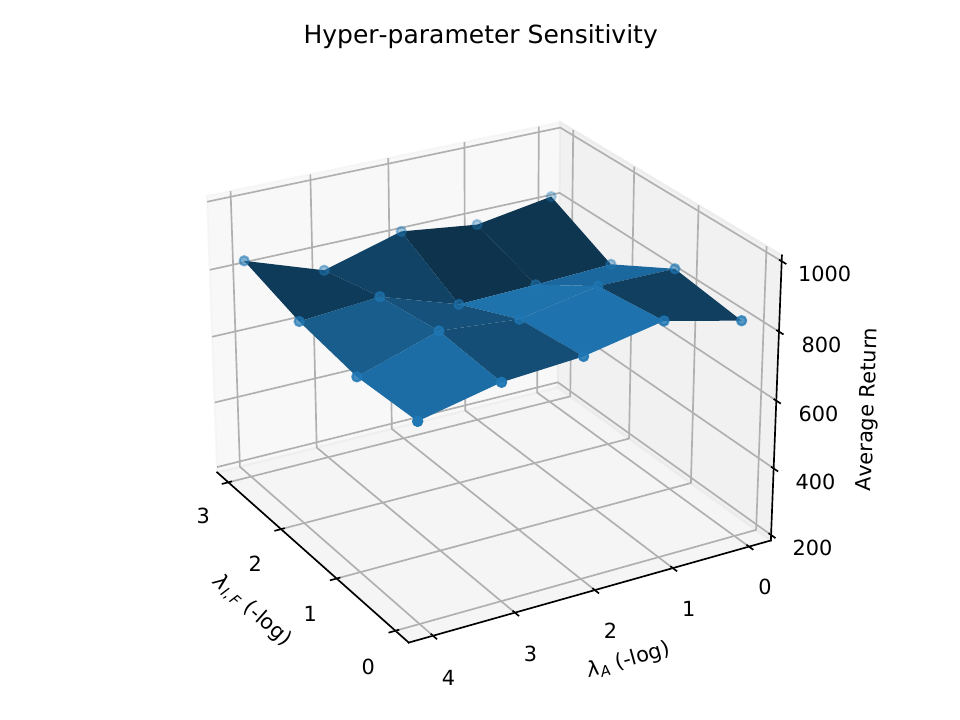}
    \vspace{-0.2cm}
\caption{Hyper-parameter sensitivity of SPD's objective weights. The value corresponding to each point is the average return over three seeds. }
\label{fig:hyper_sens}
\end{figure}
For our hyper-parameter sensitivity testing, we select $\lambda_{\psi}$ from $\{10^{-3}, 10^{-2}, 10^{-1}, 10^{0}\}$ and  $\lambda_{A}$ from $\{10^{-4}, 10^{-3}, 10^{-2}, 10^{-1}, 10^{0}\}$. 
We showed that a sensitivity evaluation on Walker Walk in the DeepMind Control suite in Figure~\ref{fig:hyper_sens}. We observe that SPD has good performance in a wide range of hyper parameter choices.
\end{document}